\documentclass[preprint, 5p, twocolumn, authoryear]{elsarticle}

\usepackage{amssymb}
\usepackage{amsmath}
\usepackage{multirow}
\usepackage{enumitem}
\usepackage{url}
\usepackage{xcolor}

\newcommand{\figureref}[1]{Figure \ref{#1}}
\newcommand{\tableref}[1]{Table \ref{#1}}
\newcommand{\equationref}[1]{Equation \ref{#1}}
\newcommand{\sectionref}[1]{Section \ref{#1}}

\newcommand{\textcite}[1]{\citet{#1}}
\newcommand{\parencite}[1]{\citep{#1}}

\journal{Ocean Engineering}

\begin{document}

\begin{frontmatter}

\title{Improving Underwater Acoustic Classification Through Learnable Gabor Filter Convolution and Attention Mechanisms}

\author[label1]{Lucas C. F. Domingos\corref{cor1}} \ead{lucas.domingos@flinders.edu.au}

\author[label1]{Russell Brinkworth} \ead{russell.brinkworth@flinders.edu.au}
\author[label2]{Paulo E. Santos} \ead{paulo.santos@priorianalytica.com}
\author[label1,label3]{Karl Sammut} \ead{karl.sammut@flinders.edu.au}

\cortext[cor1]{Corresponding Author}
\affiliation[label1]{organization={College of Science and Engineering, Flinders University},
             city={Tonsley},
             state={South Australia},
             country={Australia}}
\affiliation[label3]{organization={Centre for Defence Engineering Research and Training, Flinders University},
             state={South Australia},
             country={Australia}}
\affiliation[label2]{organization={Research And Development, PrioriAnalytica},
             city={Adelaide},
             state={South Australia},
             country={Australia}}

\begin{abstract}
Remotely detecting and classifying underwater acoustic targets is critical for environmental monitoring and defence. However, the complexity of ship-radiated and environmental noise poses significant challenges for accurate signal processing.
While recent advancements in machine learning have improved classification accuracy, limited dataset availability and a lack of standardised experimentation hinder generalisation and robustness.
This paper introduces GSE ResNeXt, a deep learning architecture integrating learnable Gabor convolutional layers with a ResNeXt backbone enhanced by squeeze-and-excitation attention.
The Gabor filters serve as two-dimensional adaptive band-pass filters, extending the feature channel representation. Its combination with channel attention improves training stability and convergence while enhancing the model’s ability to extract discriminative features.
The model is evaluated using three training–test split strategies that reflect increasingly complex classification tasks, demonstrating how systematic evaluation design addresses issues such as data leakage, temporal separation, and taxonomy.
Results show that GSE ResNeXt consistently outperforms baseline models like Xception, ResNet, and MobileNetV2, in terms of classification performance. Regarding stability and convergence, adding Gabor convolutions to the initial layers of the model reduced training time by up to 62\%. During the evaluation of training-testing splits, temporal separation between subsets significantly affected performance, proving more influential than training data volume.
These findings suggest that signal processing can enhance model reliability and generalisation under varying environmental conditions, particularly in data-limited underwater acoustic classification. Future developments should focus on mitigating environmental effects on input signals.
\end{abstract}

%%%%%%%%%%%%%%%%%%%%%%%%%%%%%%%%%%%%%%%%%%%%%%%%%%%%%%%%%%%%%%%%%%%%%%%%%%%%%%%%%%%%
%%Graphical abstract
%\begin{graphicalabstract}
%\includegraphics{grabs}
%\end{graphicalabstract}

%%%%%%%%%%%%%%%%%%%%%%%%%%%%%%%%%%%%%%%%%%%%%%%%%%%%%%%%%%%%%%%%%%%%%%%%%%%%%%%%%%%%
%%Research highlights
%\begin{highlights}
%    \item The application of learnable Gabor convolutional layers improves stability of the model during the training process, reducing training time and improving classification performance.
%    \item The incorporation of channel attention mechanisms alongside Gabor convolutional layers acts as a two-dimensional frequency selector, improving noise robustness and model convergence.
%    \item Classification performance deteriorates proportionally with increasing temporal distance between same ship recordings.
%    \item The amount of training data in ship-type acoustic classification is less influential than distance in determining classification performance.
%\end{highlights}

\begin{keyword}
Machine Learning \sep Ship-Radiated Noise \sep Spectrograms \sep Vessel Type Classification \sep Target Recognition
\end{keyword}

\end{frontmatter}

%%%%%%%%%%%%%%%%%%%%%%%%%%%%%%%%%%%%%%%%%%%%%%%%%%%%%%%%%%%%%%%%%%%%%%%%%%%%%%%%%%%%%%% main text

\section{Introduction}
\label{sec:introduction}
Remote detection of underwater acoustic targets is a valuable and fundamental asset for environmental protection and defence, as its applications include surveillance, dredging, hydrography, marine life research, and fisheries protection, to name a few \parencite{bjorno2017chapter}. Given the sensitive nature of these applications, the correct detection, processing, and classification of such signals is essential. In the defence field, for instance, incorrect classification could lead to false alarms or even to the non-identification of a potential threat. In environmental protection contexts, misclassifications can directly affect species population estimates, possibly leading to inappropriate management strategies \parencite{caillat2013effects}. 

Recently, a significant increase in anthropogenic activities conducted in the ocean has elevated the levels of underwater acoustic noise \parencite{ross2005ship, hildebrand2009anthropogenic}, driving new studies to focus on the control of maritime traffic by detecting and classifying ship types by their radiated noise. The complexity of such signals, on the other hand, poses a challenge to their effective management in certain situations. Noise recorded by passive sonars is usually composed of environmental sounds and ship-radiated noise \parencite{domingos2022survey}. Environmental sounds include sea surface noise, such as breaking waves and wind, and intermittent biological sounds, which depend on factors such as location, weather conditions, and seasonality \parencite{xu2016digital}. Ship-emitted noise, on the other hand, consists of machinery and propeller noises, which appear as constant lower components in the frequency spectra, and hydrodynamic noise, which is a broadband component that shifts between mid and high frequencies according to the ship's speed \parencite{carlton2018marine}. Additionally, multi-path sound reflection caused by interaction with different media, such as the sea surface, can affect classification accuracy within recordings \parencite{wilmut2007inversion, mckenna2012underwater}.

With the evolution of machine learning methods, their application in underwater target classification has been increasing. With a focus on feature extraction and classification of ship types, recent research has shown an interest in improving the machine learning architecture to extract relevant information from complex input data. A subset of works uses autoencoders to improve the feature extraction process \parencite{luo2020underwater, li2020feature, irfan2021deepship}, reducing the need for manual hyperparameters tuning. These architectures take advantage of the data-driven training process, which automatically extracts features from the input data, minimising a given loss function. However, the acquisition of such signals is sometimes sensitive and complex, leading to limited available data and, therefore, limited features. Another common strategy applied for detection and classification is the use of recurrent neural network blocks to extract the temporal dependency of the acoustic signal \parencite{liu2021underwater, alouani2022spatiotemporal, han2022underwater}. This strategy aims to extract features based on the temporal components of the audio; however, as the ship noise is almost constant over short periods of time, the improvements by focusing on short-time dependencies are minimal. 

Notably, the most common approach to detecting and classifying underwater acoustic signals involves an initial pre-processing stage, where the raw waveform is transformed into a spectrogram. The resulting two-dimensional representation allows the signal to be classified using algorithms originally developed for computer vision tasks. Residual neural networks (ResNet) \parencite{he2016deep} serve as the baseline for several pipelines in underwater acoustic target classification \parencite{hong2021underwater, yao2023underwater, ren2022ualf, domingos2022investigation}. They introduce identity shortcut connections between convolutional layers, enabling the network to bypass one or more layers by directly passing the output of earlier layers to later ones. These connections extend the model’s capacity by allowing the exploration of alternative paths during inference, thereby mitigating the vanishing gradient problem \parencite{veit2016residual}. Subsequent improvements introduced the concept of cardinality \parencite{xie2017aggregated}, which enhances the model’s ability to explore multiple learning paths through grouped convolution blocks within residual layers. The resulting model, ResNeXt, often achieves higher classification accuracy with only a modest increase in computational cost. In underwater acoustics, the application of ResNets has proven highly beneficial for improving classification accuracy, as they enable deeper architectures and facilitate the exploration of multiple inference paths \parencite{hong2021underwater, yao2023underwater, ren2022ualf, domingos2022investigation}. In a similar fashion, multi-branch classification approaches are also frequently employed \parencite{xie2024unraveling, xie2024advancing, wang2023underwater, tian2023joint, qian2025multi}, further enhancing the multi-path capability of the pipelines. The successful application of these strategies highlights the feasibility of employing computer vision techniques for classification tasks based on spectrograms.

Attention mechanisms have been incorporated into classification pipelines, allowing models to dynamically assign importance to input elements and focus on contextually relevant information when producing outputs~\parencite{bahdanau2016neural}. These mechanisms have been further emphasised with the advent of transformer models~\parencite{vaswani2017attention}, which rely solely on attention mechanisms and dispense with convolutions. Over the years, transformer models have been optimised for mobile applications \parencite{mehta2022mobilevit}, and variants have also been applied to spectrogram classification \parencite{gong2021ast}, highlighting their versatility in contexts outside the field of computer vision. Despite the impressive results obtained in vision applications using transformers, the combination of attention mechanisms and convolutional neural networks is still being explored, yielding superior results in several tasks~\parencite{han2023survey, jin2025effective, sheikh2025ensemble}. In vision tasks, attention is typically applied along the spatial dimension, assigning adaptive importance factors across the image~\parencite{zhu2019empirical, dosovitskiy2021image}, or along the channel dimension, prioritising different feature channels within the neural network~\parencite{wang2020ecanet, hu2018squeezeandexcitation}. Subsequent approaches also focus on combining both strategies in a spatial–channel framework~\parencite{woo2018cbam, chen2017scacnn}, which aims to enhance the model’s capability to represent and extract features from the inputs. For spectrograms, as the spatial dimension has different scales along the vertical and horizontal axes, some applications separate the time and frequency dimensions when applying spatial attention methods~\parencite{yang2024underwater}. In underwater acoustic classification, attention mechanisms have proven effective in improving accuracy by assigning different levels of importance to features at local and global scales~\parencite{zhou2023attentionbased, yang2023lightweight, xu2023underwater, yang2024underwater, wang2025adaptive}. While these techniques continue to advance classification performance, their effectiveness is highly dependent on the type of spectral representation used~\parencite{domingos2022investigation}. Processing these representations to address the complex behaviour of underwater acoustic signals offers an attractive alternative to improvements in neural networks alone.

Gabor filters, which are band-pass filters well suited to local frequency and orientation analysis \parencite{daugman1985uncertainty}, have the potential to enhance feature extraction in convolutional neural networks. Their application in underwater acoustic tasks has also been explored in recent years. In a study of whale vocalisations, Gabor filtering was applied to enhance distributed acoustic sensing (DAS) data, improving the signal-to-noise ratios of the inputs \parencite{goestchel2025enhancing}. While it supports the broader applicability of these filters in underwater environments, DAS data differs fundamentally from conventional hydrophone data, which was not addressed by this work.
In a hydrophone-based classification problem, Gabor filters were applied directly to raw audio as a preprocessing step, generating a single-channel spectral representation as input \parencite{elsborg2025acoustic}. This preprocessing stage dynamically emphasised different frequency components, improving classification performance. However, the resulting spectral input was used without further processing, leaving the task of feature extraction entirely to the model.
The application of these filters has also been studied in the two-dimensional domain, where they were applied to spectrogram inputs followed by a low-pass filter \parencite{ren2022ualf}. Despite achieving a 3.5\% improvement in accuracy, this approach focused solely on enhancing the initial representations, without further processing the resulting frequency bands. These results support the use of Gabor filters to improve generalisation, but leave open the question of how the different frequency bands can be more effectively exploited within the neural network architecture. Furthermore, most previous works consider only a single classification task, which limits the generalisation of their findings.

Despite the works produced in recent years, concerns about the generalisation of classification remain, since the availability of datasets is limited and most strategies are based on data-driven approaches. The absence of a standardised experimentation process gives rise to a range of conclusions that vary significantly depending on the chosen strategy. Additionally, optimising the feature extraction layers with a view to improving generalisation and the training stage can strongly contribute to future applications of machine learning in environments with limited amounts of data, which is often the case in underwater acoustic scenarios. In this context, the combination of enhancements to feature extraction layers in deep neural networks with improvements in the processing of those features in the network's backbone is missing from the literature, as is a standardised evaluation of the results encompassing a wide range of classification tasks.

This paper presents a method for underwater acoustic classification that optimises the feature extraction layer, providing an experimentation strategy that addresses three different scenarios and considers the variability of the training dataset. The Gabor Squeeze-and-Excitation ResNeXt model (GSE ResNeXt) introduces the usage of Gabor convolutional filters in the initial layers of the neural network combined with residual layers and channel attention mechanisms in the models' backbone, showing to be effective at improving the training stability and classification performance across different classification tasks. Additionally, the use of Gabor filters facilitated the convergence of the model, representing a substantial enhancement for future applications that utilise limited number of training examples. In summary, the contributions of this paper are as follows: 

\begin{enumerate}
    \item The integration of learnable Gabor convolutional layers with squeeze-and-excitation channel attention, serving as a two-dimensional adaptive band-pass front-end followed by frequency-selective channel weighting. This design stabilises optimisation and enhances discriminative feature learning, thereby improving classification performance.
    
    \item The application of learnable Gabor convolutions in the initial layer accelerates convergence, delivering a smoother validation trajectory and achieving steady state earlier without compromising final performance.
    
    \item A systematic evaluation of model generalisation across three training–testing split strategies of increasing complexity. This highlights the impact of temporal separation, data leakage and class taxonomy on classification performance. It also establishes a reproducible framework for benchmarking underwater acoustic classifiers.
\end{enumerate}

The fundamental background knowledge underpinning this work, including the two-dimensional Gabor convolution and the squeeze-and-excitation attention mechanism, is detailed in \sectionref{sec:concepts}, together with a description of the three classification tasks for underwater acoustics. The methods used to construct the classification pipeline, including the model architecture, are presented in \sectionref{sec:methods}. To evaluate the proposed architecture, the dataset and evaluation metric are described in \sectionref{sec:experimental}, alongside a comprehensive explanation of the experiments performed and the parameters used. The results are reported in \sectionref{sec:results}, including a comparison across the three tasks and an extensive assessment of temporal consistency in classification performance. Ablation studies are also presented in \sectionref{sec:results}, followed by the conclusions in \sectionref{sec:conclusions}.

\section{Fundamental Concepts}
\label{sec:concepts}
\subsection{Two-dimensional Gabor convolution}

Traditional convolutional blocks perform the convolution of a learnable two-dimensional kernel on an input image, occasionally adding an additional bias term depending on the application \parencite{lecun2015deep}. When visualising the convergence of convolutional networks, it was observed that the convolutional kernels in the initial layers of models trained for image classification converged to Gabor-like filters and colour blob representations \parencite{krizhevsky2012imagenet, zeiler2013visualizing}. These findings suggest that the training process could be enhanced by constraining the spatial kernels to Gabor functions, which possess significantly fewer parameters to optimise and are capable of extracting spatial textures from visual inputs. Building on these conclusions, although several studies have employed Gabor functions to pre-process and extract features from images, the integration of a Gabor filter layer with learnable weights into the back-propagation algorithm was only proposed years later \parencite{alekseev2019gabornet}. Applying this approach to image classification tasks demonstrated improved robustness, as the visual kernels were less sensitive to noisy information, while also reducing training complexity.

A Gabor function can be understood as a complex sinusoid modulated by a Gaussian envelope. While the complex sinusoid captures the frequency and phase information of the features, the Gaussian component localises them in space. In image processing, the imaginary part of the sinusoid is often omitted to simplify the calculation process, since it is sufficient for texture and feature detection in most cases \parencite{jain1991unsupervised}. In the two-dimensional domain, the real Gabor function can be defined as:

\begin{equation}
    g(x, y) = \frac{1}{2\pi\sigma^2} \exp\left( -\frac{x'^2 + \gamma^2 y'^2}{2 \sigma^2} \right) \cos\left( 2 \pi f_0 x' + \phi \right)
    \label{eq:2d_gabor}
\end{equation}

where:

\begin{align*}
    x' &= x \cos \theta + y \sin \theta \\
    y' &= -x \sin \theta + y \cos \theta .
\end{align*}

In \equationref{eq:2d_gabor}, the filter coordinates are $\left(x, y\right)$, the standard deviation of the Gaussian component and its spatial aspect ratio are given by $\sigma$ and $\gamma$ with the filter oriented at an angle $\theta$. For the sinusoidal wave, the fundamental frequency is given by $f_0$ and the phase by $\phi$.

Given that the frequency response of a Gabor filter is a shifted Gaussian, its band-pass nature becomes particularly useful for texture and line extraction in spatial representations. Additionally, its monotonicity and differentiability facilitate its application with back-propagation algorithms. By learning only the filter parameters, the application of a Gabor convolutional layer simplifies the learning process and provides robust initialisation to the model. 

\subsection{Squeeze-and-Excitation attention}

In a neural network, convolutional blocks are feature extraction layers that fuse spatial and channel-wise information into a set of representations. Despite the good generalisation achieved by these blocks, there is a constant effort to improve their performance on specific tasks, adapting them for target applications. The Squeeze-and-Excitation (SE) \parencite{hu2018squeezeandexcitation} block emerges as an alternative solution to this problem, modelling the channels interdependency on the convolutional features. The SE block can be understood as a channel-wise attention mechanism that aggregates and recalibrates the channel-wise feature responses. 

The initial stage of this block is the squeeze operation, which involves aggregating channel features into a channel descriptor. By applying a Global Average Pooling operation, the compression of the spatial dimensions $H \times W$ into a channel descriptor $z_c$ for the \emph{c-th} channel is described as:

\begin{equation}
    z_c = \frac{1}{H \times W} \sum_{i=1}^{H} \sum_{j=1}^{W} x_c(i,j).
    \label{eq:squeeze}
\end{equation}

The channel descriptors act as an embedded representation of the feature channels. Once obtained, an adaptive gating operation is performed by applying a linear layer with learnable parameters and a nonlinear activation function to extract useful relationships from those embeddings, resulting in a scaling factor $s$. Considering $\sigma$ as the sigmoid function, $\delta$ as the ReLU activation and $\Theta$ as the weights of the linear model, given a channel descriptor $z$, the scaling factor is described by the following operation:

\begin{equation}
    s= \sigma \left( \Theta_2 \delta \left( \Theta_1 z\right)\right)
\end{equation}

With a slight increase in computational cost, the \emph{squeeze-and-excitation} operation applied to the ImageNet dataset achieved a significant improvement of nearly 25\% over previous models \parencite{hu2018squeezeandexcitation}. When applied to spectral representations, it enhances channel-wise information by capturing nonlinear relationships between multiple two-dimensional features. By combining both channel attention and Gabor convolutional layers within a neural network architecture, it becomes possible to adaptively select two-dimensional frequency bands. 

\subsection{Classification tasks in underwater acoustics}
\label{subsec:tasks}

In underwater acoustic classification using machine learning frameworks, several combinations of hyper-parameters are found in the literature. They range from audio segment duration and number of classes to the metric calculation process, making direct comparisons of such works difficult and sometimes inappropriate. In addition, some results are often incompatible due to the various manners in which the training and test datasets are divided. Considering this, three classification tasks have been proposed \parencite{liu2021underwater}: \emph{Task 1} consists of first segmenting the original audio into smaller parts, randomising these segments, and dividing them further into training and test subsets; \emph{Task 2} utilises the beginning and end of each audio file for training or testing, and then segments it into smaller parts; and \emph{Task 3} consists of using each of the entire audios as either a training or test sample, then segmenting it into smaller parts. The tasks gradually increase in difficulty, providing a broader view of the classification problem.

The aforementioned tasks have different characteristics and limitations, which, when considered together, can provide a more comprehensive understanding of the classification problem. \emph{Task 1} has been widely studied in the literature, with some studies achieving accuracy rates of around 99\%. Despite these results, \emph{Task 1} is far from the real-world problem, and its high-performance rates are closely related to the models' ability to store information (i.e. to overfit) rather than to generalise it. As the training and test data are generated from interleaved audio, they share similarities that facilitate classification but also make the models prone to overfitting. 

\emph{Task 2} provides insights into two major concerns: the fluctuation in classification performance as the ship signal changes over time, and the ability to extract sufficient information from a short audio recording to enable the detection of the same target at a later time. In this context, \emph{Task 2} presents a more realistic scenario than \emph{Task 1}, as it increases the separation between training and test datasets while maintaining subsets related to the same target. The primary concern associated with \emph{Task 2} lies in the generalisation of the results in a classification setting. Since different segments of the same recordings are present in both subsets, background noise and environmental conditions are also shared, raising the question of whether the results are applicable to other conditions. \emph{Task 3} is introduced as a mitigation strategy to complement the conclusions drawn from \emph{Task 2} under more diverse conditions.

Similarly to \emph{Task 1}, \emph{Task 3} is widely covered in the specialised literature, however, its performance still has some limitations when approaching real-world applications. These limitations can be exacerbated by its application in ship-type classification since different ship types do not have unique characteristics that can distinguish between them completely, resulting in overlap between classes and hindering accuracy. This also raises the question of whether other characteristics should be used as targets instead of vessel type alone (e.g. number of propellers, vessel size, engine type). Regarding the classes overlap, the advantage of analysing the \emph{Task 2} results is that they provide complementary information showing whether there is a clear separation between classes, even when the vessels are the same. \emph{Task 3} tries to generalise this separation for different vessels. This work considers all three tasks as evaluation strategy for the classification pipelines.

\section{Methods}
\label{sec:methods}
\subsection{Gabor Squeeze-and-Excitation ResNeXt}

This works proposes a novel Gabor Squeeze-and-Excitation ResNeXt (GSE ResNeXt) model, which incorporates residual connections, grouped convolutions, and SE attention modules into a model architecture, also replacing the first convolutional block with a Gabor convolution. This model combines the stability and optimisation provided by the Gabor filters, which enhance initial layer representations, with parallel processing obtained through increased cardinality and channel-wise scaling of information. Assuming that each Gabor filter kernel extracts distinct frequency band information, the channel attention blocks act as a gating mechanism, improving the relative importance of two-dimensional frequencies. 

During the training procedure, the initialisation of the Gabor function parameters followed the strategy proposed in \parencite{alekseev2019gabornet, meshgini2012face}. Considering \equationref{eq:2d_gabor} as the Gabor kernel, the aspect ratio $\gamma$ was set to 1, and the relationship between $\sigma$ and $\omega$, which is the angular frequency ($\omega = 2 \pi f$), was initialised as $\sigma = \pi / \omega$. The frequencies and orientations of the kernels are initialised using the following equations:

\begin{align}
    \omega_m = \frac{\pi}{2}\sqrt{2}^{-m} & , & \theta_n = \frac{\pi}{8}\left(n-1\right) 
    \label{eq:gabor_freq_ori}
\end{align}

where:

\begin{align*}
    m &= 1,2,...,5 \\
    n &= 1,2,...,8.
\end{align*}

In spectral representations, where only one channel is input to the network, the usage of Gabor convolutions on the initial layer can increase the network's robustness, as the kernels are already constrained to visually meaningful representations. This can also facilitate the convergence, since only the Gabor parameters need to be learnt rather than the entire kernel. As its frequency response is equivalent to that of a band-pass filter, applying multiple Gabor functions with different frequencies and orientations can expand the original input representation into multiple channels. These channels can be adaptively selected using channel attention mechanisms, such as the squeeze-and-excitation attention block.

As the spectrographic information is single-channel and the training datasets are limited in size, an 26-layer architecture was adopted for the experiments in this study. \tableref{tab:resnext26} presents the layer-by-layer design of the GSE ResNeXt model. Although the 26-layer configuration was selected for this specific setup, deeper architectures remain viable for other data-driven applications.

%\begin{table}[ht]
%\centering
%    \begin{tabular}{lcc}
%    \hline
%    \textbf{Stage} & \textbf{Building Block} & \textbf{Output Size} \\ \hline
%    Gabor Conv1 & $7 \times 7$, 64, stride 2 & $112 \times 112$ \\
%     & $3 \times 3$ max pool, stride 2 & $56 \times 56$ \\ \hline
%    Conv2 & $\begin{bmatrix}3 \times 3, 64, \textit{C}=32 \\ 3 \times 3, 64, \textit{C}=32 \\ \text{SE attention} \end{bmatrix} \times 2$ & $56 \times 56$ \\ \hline
%    Conv3 & $\begin{bmatrix}3 \times 3, 128, \textit{C}=32 \\ 3 \times 3, 128, \textit{C}=32 \\ \text{SE attention} \end{bmatrix} \times 2$ & $28 \times 28$ \\ \hline
%    Conv4 & $\begin{bmatrix}3 \times 3, 256, \textit{C}=32 \\ 3 \times 3, 256, \textit{C}=32 \\ \text{SE attention} \end{bmatrix} \times 2$ & $14 \times 14$ \\ \hline
%    Conv5 & $\begin{bmatrix}3 \times 3, 512, \textit{C}=32 \\ 3 \times 3, 512, \textit{C}=32 \\ \text{SE attention} \end{bmatrix} \times 2$ & $7 \times 7$ \\ \hline
%     & Global average pool, & \multirow{2}[1]{*}{$1 \times 1$} \\
%     & 4-d fc, softmax &  \\ \hline
%    \end{tabular}
%\caption{The GSE ResNeXt-18 architecture. \textit{C=32} denotes the cardinality of the grouped convolutions. The first layer, \textit{Gabor Conv1}, represents the learnable Gabor convolution. Table inspired by the one provided in \textcite{xie2017aggregated}.}
%\label{tab:resnext18}
%\end{table}

\begin{table}[ht]
\centering
\setlength{\tabcolsep}{4pt}
\resizebox{\columnwidth}{!}{
    \begin{tabular}{lcc}
    \hline
    \textbf{Stage} & \textbf{Building Block} & \textbf{Output Size} \\ \hline
    Gabor Conv1 & $7 \times 7$, 64, stride 2 & $128 \times 128$ \\
     & $3 \times 3$ max pool, stride 2 & $64 \times 64$ \\ \hline
    Conv2 & $\begin{bmatrix}1 \times 1, 128 \\ 3 \times 3, 128, \textit{C}=32 \\ 1 \times 1, 256 \\ \text{SE attention} \end{bmatrix} \times 2$ & $64 \times 64$ \\ \hline
    Conv3 & $\begin{bmatrix}1 \times 1, 256 \\ 3 \times 3, 256, \textit{C}=32 \\ 1 \times 1, 512 \\ \text{SE attention} \end{bmatrix} \times 2$ & $32 \times 32$ \\ \hline
    Conv4 & $\begin{bmatrix} 1 \times 1, 512 \\ 3 \times 3, 512, \textit{C}=32 \\ 1 \times 1, 1024 \\ \text{SE attention} \end{bmatrix} \times 2$ & $16 \times 16$ \\ \hline
    Conv5 & $\begin{bmatrix}1 \times 1, 1024 \\ 3 \times 3, 1024, \textit{C}=32 \\ 1 \times 1, 2048 \\ \text{SE attention} \end{bmatrix} \times 2$ & $8 \times 8$ \\ \hline
     & Global average pool, & \multirow{2}[1]{*}{$1 \times 1$} \\
     & 4-d fc, softmax &  \\ \hline
    \end{tabular}
}
\caption{The GSE ResNeXt architecture. \textit{C=32} denotes the cardinality of the grouped convolutions. The first layer, \textit{Gabor Conv1}, represents the learnable Gabor convolution. Table inspired by the one provided in \textcite{xie2017aggregated}.}
\label{tab:resnext26}
\end{table}

\section{Experimental Setup}
\label{sec:experimental}
\subsection{Datasets}
In the context of radiated ship noise, the DeepShip database \parencite{irfan2021deepship} is widely used. It is derived from the Ocean Networks Canada Society database,\footnote{\url{https://oceannetworks.ca}} which contains unlabelled underwater acoustic data. DeepShip recordings were captured between May 2016 and October 2018 at the Strait of Georgia delta node and are segmented into three time windows with hydrophones positioned at depths of 141 m, 147 m, and 144 m. Data acquisition was performed using an icListen smart hydrophone offering a frequency range of 1 Hz to 12 kHz and a sampling rate of 32,000 Hz. Vessel labels were assigned based on Automatic Identification System (AIS) data and include four categories: tugboat, passenger ship, cargo ship, and oil tanker. To ensure clean samples, only recordings containing signals from a single vessel within a 2 km radius were retained. These recordings were truncated and discarded once a second ship entered this range. The final dataset consisted of 613 recordings from 265 distinct vessels, totalling approximately 47 hours. It also provides information on the vessel’s distance from the hydrophone at three different time points to track ship trajectories. While \parencite{irfan2021deepship} proposes a five-class scheme including background noise, only the four vessel categories are present in the dataset. Some studies address this by sourcing background noise recordings from other datasets. In this study, however, only the four provided classes are used, without the inclusion of background noise recordings. 

\subsection{Classification Metrics}

In a classification pipeline, model performance is typically evaluated based on overall accuracy, defined as the ratio of correct classifications out of the total number of instances. However, accuracy alone can lead to misinterpretation of the results, since it does not distinguish between true positives ($tp$), true negatives ($tn$), false positives ($fp$) and false negatives ($fn$). Metrics such as Precision ($\frac{tp}{tp + fp}$), Recall ($\frac{tp}{tp + fn}$), and the F1-score, which is the weighted harmonic mean of precision and recall, usually provide a global intuition about the results. However, the Matthews Correlation Coefficient (MCC) \parencite{matthews1975comparison} is usually a more appropriate metric \parencite{domingos2022survey}, as it assigns a high score to predictions that are correct for all four base categories ($tp$, $fp$, $tn$ and $fp$). This coefficient is defined by the following expression:

\begin{multline}
\text{MCC} = \\ \frac{tn\times tp - fn\times fp}{\sqrt{(tp + fp)\times(tp + fn)\times(tn + fp)\times(tn + fn) }}.
\end{multline}

While most studies in the literature use micro-averaged accuracy to report their conclusions, there are several limitations to relying on accuracy alone. For example, in this four-class classification problem, a random predictor would achieve approximately 25\% accuracy, whereas the same predictor would yield a MCC of 0\%, correctly indicating randomness. To minimise the impact of data imbalance on the results, MCC is used as the reference metric, complementing the accuracy information.

\subsection{Framework and Hyper-parameters}

To encompass the most common classification strategies, the performance of the GSE ResNeXt model was compared with that of several well-established models in the literature: a classic residual network (ResNet) \parencite{he2016deep}, serving as a baseline to assess the improvements introduced in this study over the traditional architecture; MobileNetV2 \parencite{sandler2018mobilenetv2}, an efficient lightweight CNN that employs inverted residuals with linear bottlenecks, enabling analysis of the relationship between model compactness and accuracy; Xception \parencite{chollet2017xception}, an architecture that employs multi-scale residual blocks and depthwise separable convolutions, allowing comparison with the multi-path feature extraction strategy; MobileViT \parencite{mehta2022mobilevit}, which combines convolutional layers with vision transformers to balance local and global feature representations; and AST (Audio Spectrogram Transformer) \parencite{gong2021ast}, a transformer-based model designed for audio spectrogram inputs, enabling evaluation of attention-driven architectures in underwater acoustic classification.

To emphasise the lower-frequency components of ship-radiated noise, Constant Q Transform (CQT) spectrograms \parencite{brown1991calculation} were selected as input, following conclusions drawn in recent literature \parencite{domingos2022investigation, irfan2021deepship}. The CQT enhances low-frequency resolution using geometrically spaced filters and a constant quality factor, making it well-suited for identifying harmonic structures in underwater acoustic signals. The CQT hyperparameters were selected based on the characteristics of the audio signals in the DeepShip dataset, as shown in \tableref{table:cqt_preprocessing}.
The choice of using a sample rate of 32,000 samples per second ensures that the full spectral content of the signal is preserved. This allows the feature extraction process to select relevant information during training rather than constraining the input to a predefined range.
Audio segments of 30 seconds in duration were used, with the SpecAugment \parencite{park2019specaugment} strategy applied as a data augmentation step to improve training robustness, in line with recent studies \parencite{tian2023joint, hong2021underwater, liu2021underwater, zhou2023attentionbased}. Training was conducted using the AdamW optimiser with a learning rate of $1\times10^{-3}$ and a weight decay of $1\times10^{-4}$ for regularisation, over 50 epochs with a mini-batch size of 64. Cross-entropy loss with label smoothing was used as criterion during the training procedure.

\begin{table}[!ht]
\centering
	\begin{tabular}{ll}
	\hline
	\textbf{Parameter} & \textbf{Value}\\
	\hline
        Sample Rate & 32000 samples/s\\
        Duration & 30 s\\
        Overlap & 15 s\\
        Hop Length & 25 ms\\
        Minimum Frequency & 10 Hz\\
        Maximum Frequency & 16 kHz\\
        Number of bins & 255 bins\\
        Bins per Octave &  24 bins\\
	\hline
	\end{tabular}
\caption{Parameters applied to obtain the CQT spectrogram.}
\label{table:cqt_preprocessing}
\end{table}

To avoid biased conclusions derived from the dataset division, a stratified 5-fold cross-validation scheme was performed. In this scheme, four folds were used for training and the remaining one was used for testing. Training was then repeated once for each configuration, resulting in five pieces of training that were averaged to achieve the final result. For \emph{Task1}, all the 30-second audio segments were randomised and included in one of the folds, respecting the proportion of classes per fold. Considering \emph{Task 3}, each one of the entire audio files was included in a different fold and then was segmented into 30-second parts. 

As \emph{Task 2} preserves the temporal consistency of the audio files, a more complete set of test configurations was established in the analysis. Each audio file was divided into five equal-length segments, each of which was included in a different fold. After this division, they were segmented into 30-second pieces for spectrogram generation. Thus, Fold 1 contains the beginning of each audio file, Fold 5 contains the end, and Folds 2, 3, and 4 contain the intermediate parts. For a complete analysis of the results, two folds were considered for training: the initial segment of the audio (Fold 1) and the ending (Fold 5). The results were then analysed using the remaining segments as the testing subset, producing only two combinations: training on the first fold and testing on the last four, and training on the last fold and testing on the first four. As at least five 30-second segments were needed for \emph{Task 2}, and to maintain consistency with other tasks, only audio files with a duration of more than one minute and 30 seconds were considered for all the three tasks. The total number of 30-second audio segments per class and their respective proportions in the dataset are shown in \tableref{tab:class_distribution}, which also indicates that class balance was maintained.

\begin{table*}[!ht]
\centering
    \begin{tabular}{lcccc}
    \hline
     & \textbf{Tug} & \textbf{Tanker} & \textbf{Cargo} & \textbf{Passenger Ship} \\
    \hline
    \textbf{Number of Segments}       & 2668 & 2647 & 2519 & 2905 \\
    \textbf{Proportion (\%)}  & 24.85\% & 24.64\% & 23.46\% & 27.06\% \\
    \hline
    \end{tabular}
\caption{The dataset class distribution including the number of 30 second segments and its proportion of total dataset.}
\label{tab:class_distribution}
\end{table*}

To evaluate the model’s ability to separate classes across all three tasks, an analysis based on latent features was conducted. The feature representations obtained before the classification layer were extracted using the test subset and projected into two dimensions using the UMAP algorithm \parencite{mcinnes2020umap}. A comparison was made between the latent space representations categorised by the true labels in the dataset and those categorised according to the model’s predictions. These outcomes provide insight into the model’s capacity to extract meaningful features from the training data and generalise effectively to the test data. It also gives insight on the isolation between training and test subsets.

Since the dataset comprises cropped parts of the original audios, additional experiments were also performed for \emph{Task 2}, aiming to provide insights into the temporal dependency of the classification, evaluating if training and testing on adjacent segments will result in higher classification accuracy when compared with segments further apart. Additionally, the dataset size was analysed with respect to its importance for classification. Three different experiments were conducted in this context:

\begin{enumerate}[label=(\alph*)]
    \item The first experiment considered the beginning of the audio as the test subset, using one training subset at a time and evaluating how the classification performance was affected by the temporal distance between subsets.
    \item The second experiment focused on the amount of data needed to train the model. It used a fixed subset for testing and adjacent subsets of decreasing size for training. The purpose was to evaluate how reducing the size of the training dataset affected classification performance when the test and training subsets were neighbours.
    \item The third experiment focused on both temporal proximity and the amount of training data. It evaluated how the classification performance changed when subsets in the temporal domain changed in combination with dataset size.
\end{enumerate}

\begin{figure}[!bht]
    \centering
    \resizebox{\columnwidth}{!}{\includegraphics{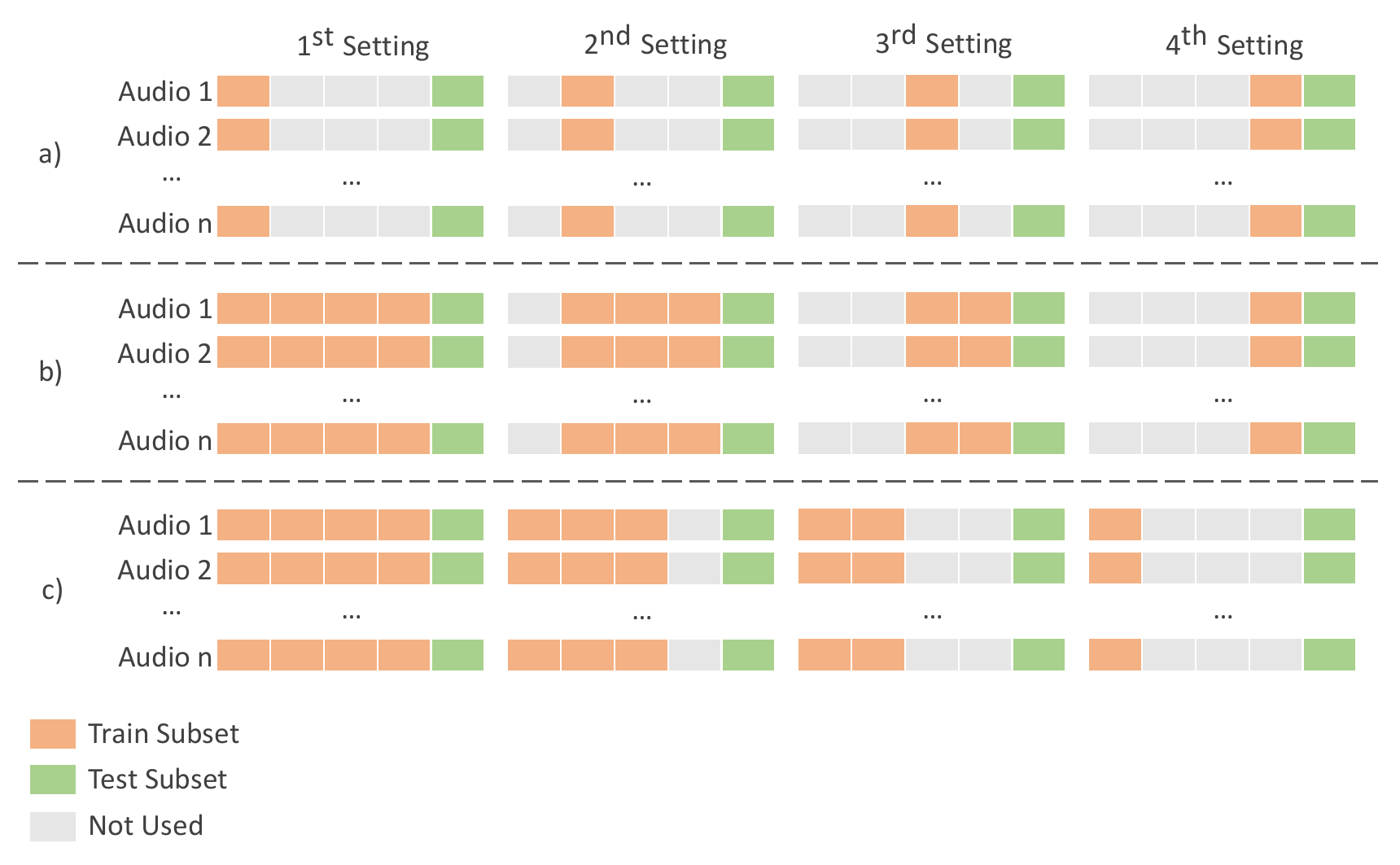}}
    \caption{Illustration of the three different experiments performed for Task 2. Experiment a) explores the temporal proximity, b) explores data reduction and c) is the combination of both.} 
    \label{fig:task2_exp}
\end{figure}

An illustration representing the three experimental scenarios is presented in \figureref{fig:task2_exp}. To provide a comprehensive understanding of the results, the analogous counterpart of each setting was also considered in the experiments by mirroring the folds used for training and testing, i.e. using the first audio segment as the test subset instead of the final one as seen in \figureref{fig:task2_exp}. The results consider the average value of these two configurations.

\section{Results and Discussion}
\label{sec:results}

Experiments were conducted across all configurations to measure baseline MCC values for the three tasks, as shown in \tableref{table:baseline_tasks}. The model size was quantified by parameter count, and computational complexity was assessed via floating-point operations (FLOPs).

\begin{table*}[tb]
\centering
\setlength{\tabcolsep}{4pt}
\resizebox{\textwidth}{!}{
	\begin{tabular}{lcccccccc}
	\hline
        \multirow{2}[1]{*}{\textbf{Model}} & \multirow{2}[1]{*}{\textbf{Params}} & \multirow{2}[1]{*}{\textbf{FLOP}} & \multicolumn{3}{c}{\textbf{MCC (\%)}} & \multicolumn{3}{c}{\textbf{Accuracy (\%)}}\\
	   & & & \textbf{Task 1} & \textbf{Task 2} & \textbf{Task 3} & \textbf{Task 1} & \textbf{Task 2} & \textbf{Task 3}\\
	\hline
	MobileNet V2 & 2.2M  & 0.4G & 99.19 $\pm$ 0.15 & 65.12 $\pm$ 2.35 & 59.08 $\pm$ 6.22 & 99.39 $\pm$ 0.11 & 73.82 $\pm$ 1.78 & 69.01 $\pm$ 4.93 \\
	ResNet 18    & 11.2M & 2.1G & 99.42 $\pm$ 0.33 & 75.34 $\pm$ 2.34 & 64.15 $\pm$ 3.64 & 99.56 $\pm$ 0.25 & 81.46 $\pm$ 1.79 & 72.75 $\pm$ 3.06 \\
	Xception     & 20.8M & 5.6G & \textbf{99.63 $\pm$ 0.14} & 76.55 $\pm$ 0.89 & 63.00 $\pm$ 3.29 & \textbf{99.72 $\pm$ 0.10} & 82.39 $\pm$ 0.66 & 72.03 $\pm$ 2.72 \\
    AST          & 5.9M  & 7.8G  & 92.61 $\pm$ 1.42 & 59.33 $\pm$ 0.53 & 55.05 $\pm$ 6.01 & 94.45 $\pm$ 1.07 & 69.48 $\pm$ 0.43 & 66.10 $\pm$ 4.71 \\
    MobileViT    & 5.0M  & 1.7G  & 98.65 $\pm$ 0.13 & 65.51 $\pm$ 0.71 & 57.10 $\pm$ 5.78 & 98.99 $\pm$ 0.09 & 74.05 $\pm$ 0.52 & 67.66 $\pm$ 4.48 \\
	\textbf{GSE ResNeXt} & 14.7M & 3.0G & 99.53 $\pm$ 0.15 & \textbf{78.53 $\pm$ 2.50} & \textbf{64.86 $\pm$ 4.83} & 99.65 $\pm$ 0.11 & \textbf{83.85 $\pm$ 1.85} & \textbf{73.31 $\pm$ 3.75} \\
	\hline
	\end{tabular}
}
\caption{Baseline results obtained for the three classification tasks. Data is provided as mean $\pm$ standard deviation across folds, and accuracy is presented as complementary information alongside MCC.}
\label{table:baseline_tasks}
\end{table*}

Analysis of the results presented in \tableref{table:baseline_tasks} shows that \emph{Task 1} achieves an MCC of around 99\% for four out of the six models, showing that the similarity between adjacent audio segments is sufficient for the model to extract almost 100\% of the classification information from the training dataset. Variations in model size and architecture have minor interference with the final classification results, as the models are probably only fitting the training data and not providing enough information to evaluate their application to unseen data. While these results demonstrate that the input data can be classified using representative data, they also highlight the need for a more comprehensive evaluation of the approach, focusing on tasks 2 and 3.

\begin{figure}[!bht]
    \centering
    \resizebox{\columnwidth}{!}{\includegraphics{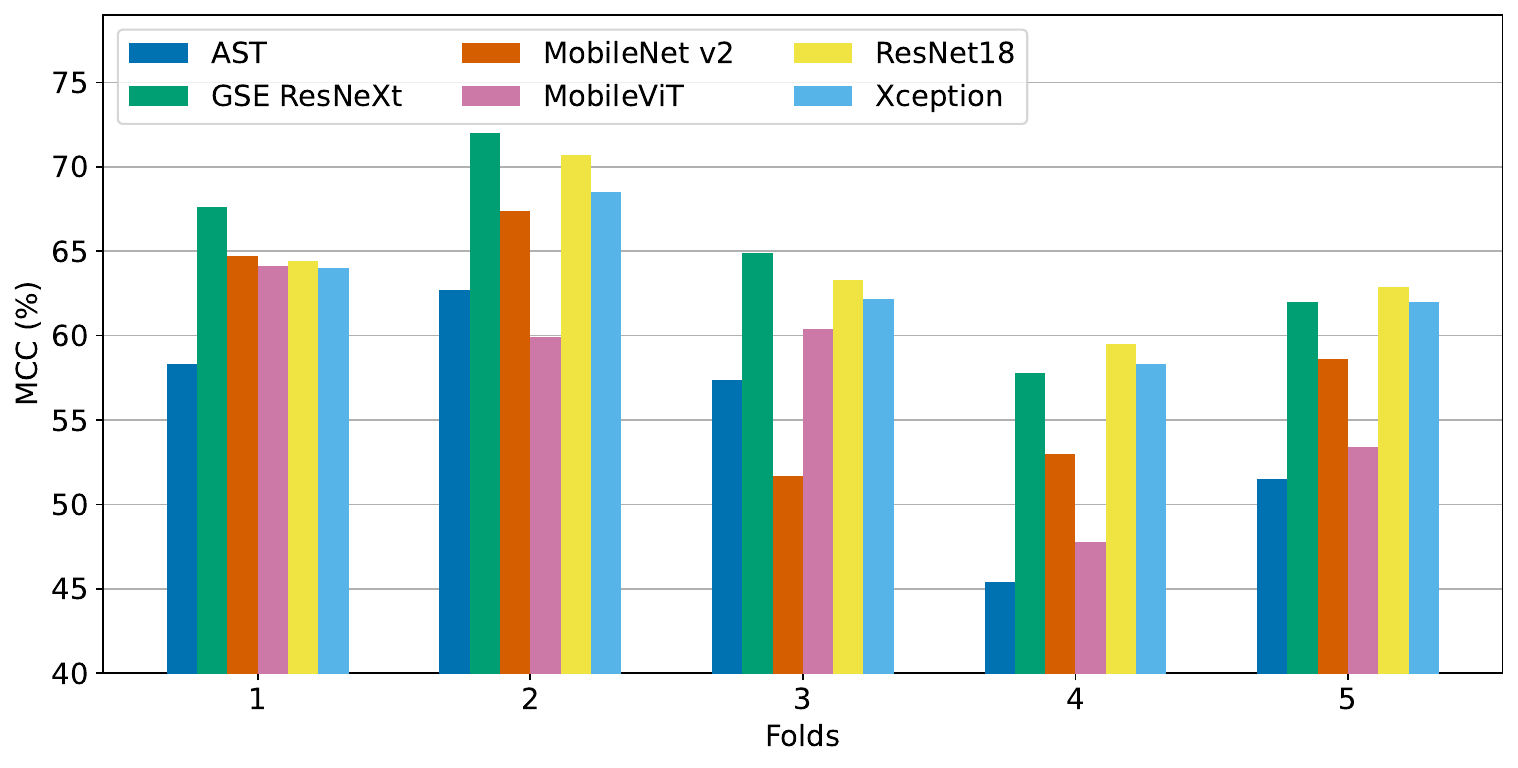}}
    \caption{The classification performance (MCC) in \emph{Task 3} obtained for each model across all the 5 different folds. Classification performance varied substantially between folds, leading to a higher standard deviation.}
    \label{fig:task3_folds}
\end{figure}

The results obtained for \emph{Task 3} show an accuracy drop of approximately 35 MCC percentage points compared to \emph{Task 1}, reflecting the improved isolation between the training and test subsets. With reduced similarities between subsets, since the same ship recording was not present in both, the classification task becomes more complex, resulting in a significant decrease in performance. This task also exhibited greater variability in performance across different folds, with a standard deviation exceeding six percentage points when using MobileNet V2 and AST, two of the worst-performing models. This variation raises concerns regarding the similarities between different ship classes and their taxonomy, since classification results differ significantly depending on the folds division, as illustrated in \figureref{fig:task3_folds}. Since there are different recordings in each fold, radiated noise of the same ship is unlikely to be placed on both the training and test datasets, making the accurate classification of such signals challenging.

\begin{figure}[!bht]
    \centering
    \resizebox{\columnwidth}{!}{\includegraphics{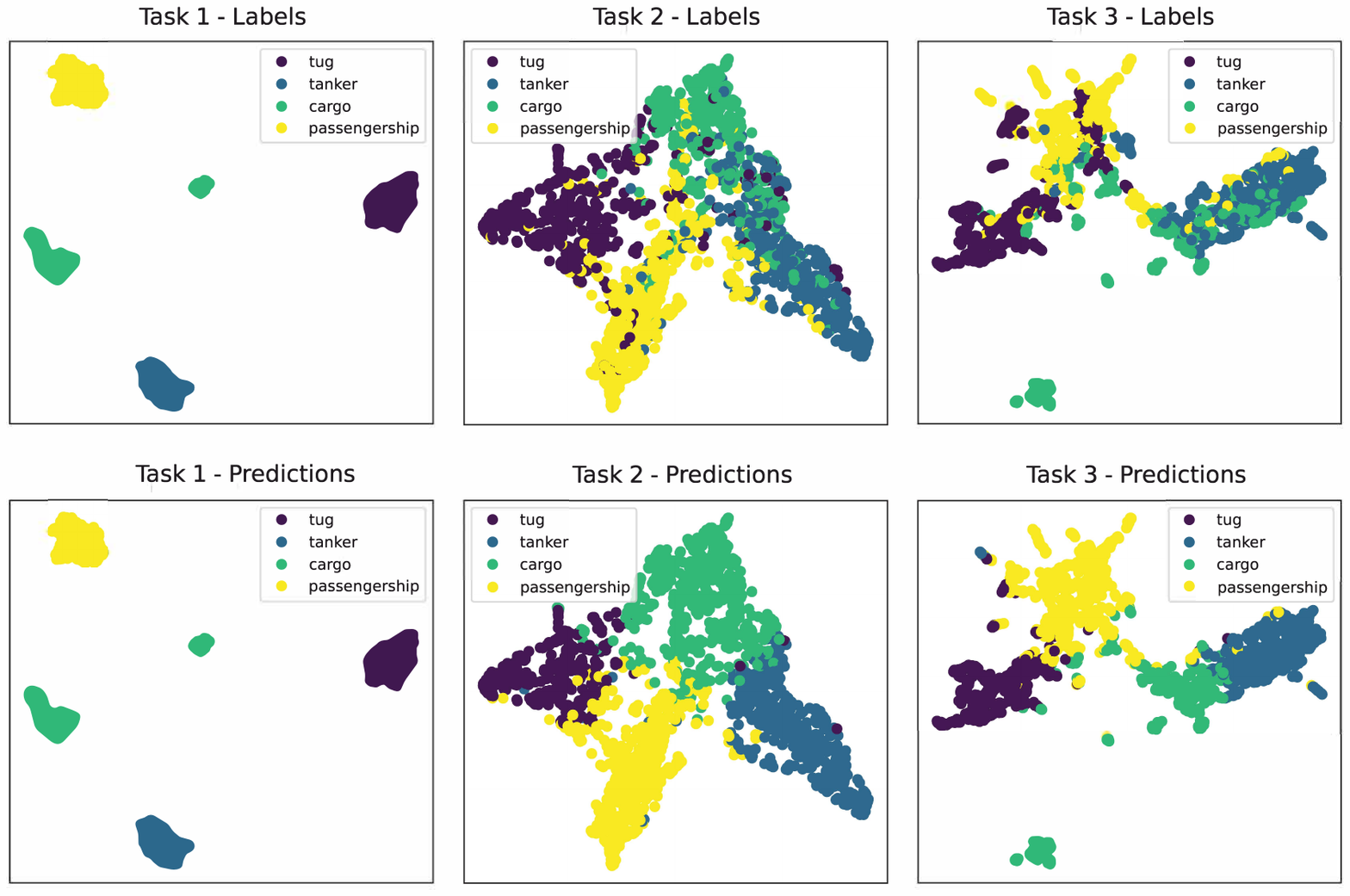}}
    \caption{The two-dimensional visualisation of the features obtained after the feature extraction stage in the GSE ResNeXt model for the three tasks considered. The top row displays the ground truth labels, while the bottom row shows the predicted labels. While the latent space for \emph{Task 1} exhibits clearer separation between classes, \emph{Task 3} reveals a more complex structure, even in the ground truth labels. \emph{Task 2} presents a marginally separable space, with minor overlaps between classes.}
    \label{fig:tasks_latent}
\end{figure}

\emph{Task 3} results also suggest that the taxonomy of classes based on ship types is unrepresentative, as the classes may not be clearly separated. To evaluate this hypothesis, the analysis based on latent features was performed using the GSE ResNeXt model, as shown in \figureref{fig:tasks_latent}. The top row shows the latent space representations categorised by the true labels in the dataset, while the bottom row shows the same representations categorised according to the model’s predictions. The difference between the tasks is noticeable, with the class boundaries in \emph{Task 1} being extremely well-defined, whereas those for \emph{Task 3} are mixed into a complex shape. Since \emph{Task 1} includes similar portions of the same recording in both the training and test datasets, the separation of the features is well-defined, producing a model that can distinguish between classes based on their training instances. \emph{Task 3}, on the other hand, evaluates the inverse relationship, due to the complete separation of entire ship recordings it focuses on model performance in the absence of data leakage between the training and test sets. The complex boundaries obtained for the test dataset reflect this characteristic. Although the features were separated during training, those extracted from different classes were similar enough to be placed in sometimes overlapping coordinates in the latent space during testing. The conclusions from this analysis point in two directions. First, the training subset alone cannot represent the entire distribution of unseen data, meaning that even if the model achieves 100\% accuracy on the training set, it will not generalise well enough to data outside that distribution. Second, there is no ideal way to categorise ship recordings into clearly defined classes, which tend to overlap in practice (e.g., a cargo ship and a tanker may sound nearly identical). A low-level taxonomy focusing on the physical characteristics of ships may offer a better solution in this case.

\begin{figure}[!bht]
    \centering
    \resizebox{0.8\columnwidth}{!}{\includegraphics{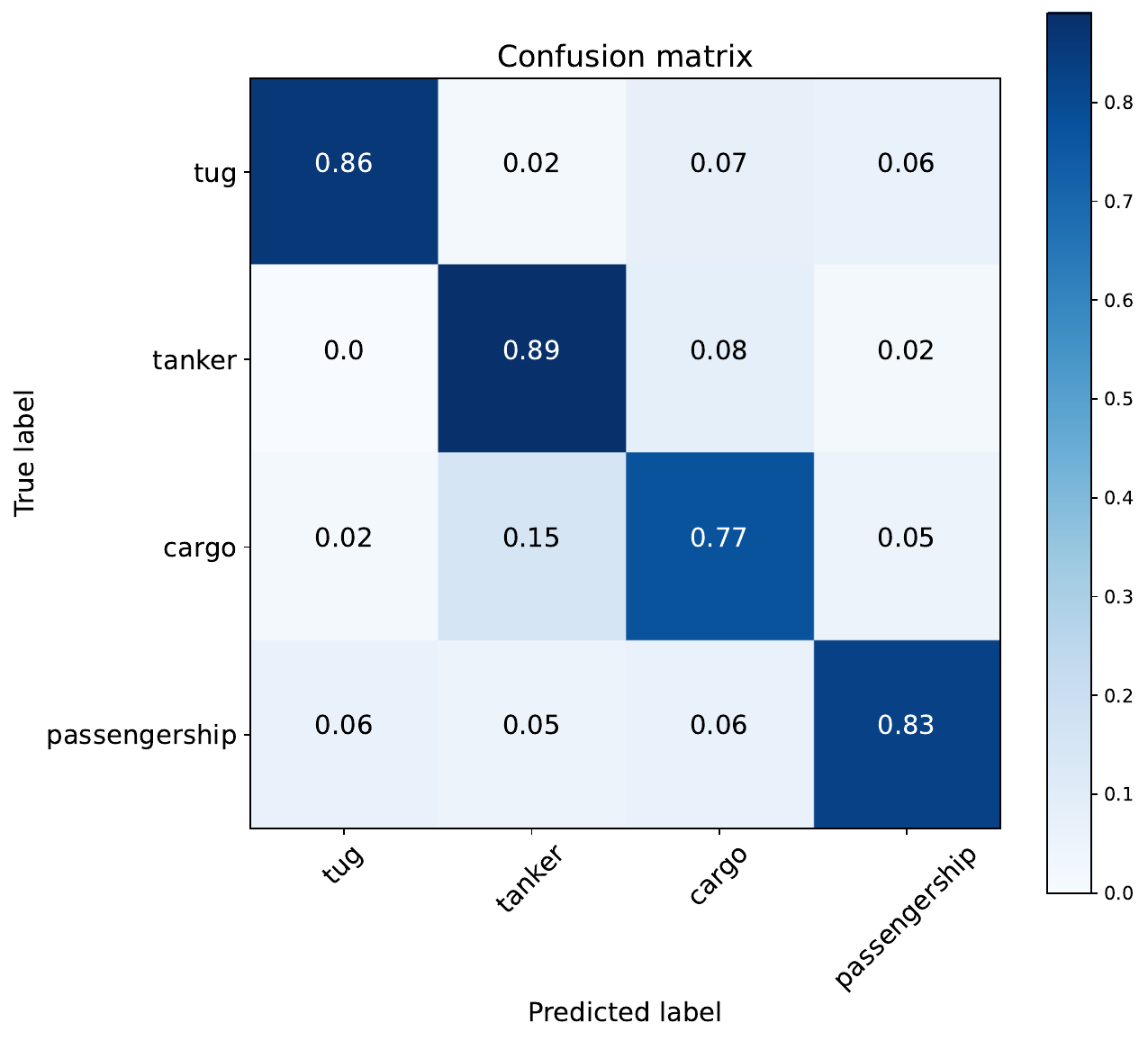}}
    \caption{Confusion matrix obtained for \emph{Task 2} using the GSE ResNeXt model. The matrix shows the distribution of correct predictions and misclassifications across ship classes.}
    \label{fig:confusion_matrix}
\end{figure}

The analysis of \emph{Task 2} aims to address the open discussions about the lack of distinctiveness between the  group classifications from \emph{Task 3}. Since the training and test datasets contain the same recordings, the taxonomy of classes can be better explored. Additionally, using consecutive audio segments, rather than the overlapping ones from \emph{Task 1}, mitigates the data leakage problem.
The classification results in \tableref{table:baseline_tasks} are complemented by the confusion matrix for \emph{Task 2} using the GSE ResNeXt model in \figureref{fig:confusion_matrix}, which highlights the distribution of correct and misclassified samples across classes. \emph{Task 2} achieved accuracy levels between those observed in \emph{Tasks 1} and \emph{3}.
The latent space representations obtained for this task are also intermediate between those of \emph{Tasks 1} and \emph{3}, exhibiting a moderate level of class separation with some overlap at the boundaries, while still maintaining a complex overall structure. The middle region of the latent space shows intersection of classes, suggesting the similar conclusions obtained for \emph{Task 3} about the taxonomy. The smaller standard deviation between folds also suggest that there is a smaller impact of the folds division in the final result. It points out that when using the same recordings, thereby the same ships, for training and test, the results are more consistent, therefore the distributions between the subsets are similar even without data leakage. This consistency is fundamental to compare the results obtained on different pipelines, alleviating dataset biases in the classification results.

\begin{figure*}[!bht]
    \centering
    \resizebox{\textwidth}{!}{\includegraphics{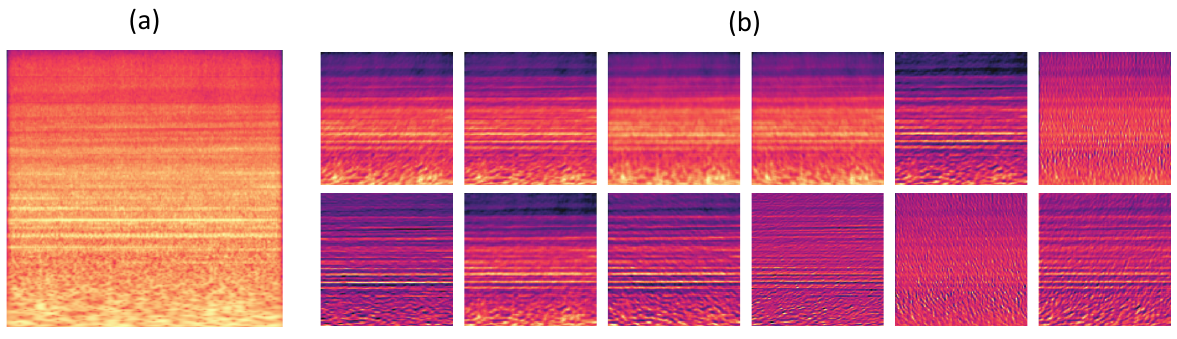}}
    \caption{An example of input spectrogram (a) and the 12 feature maps with the highest activation peaks extracted from the Gabor convolution layer of the GSE ResNeXt model (b). Maps are ordered by activation magnitude, illustrating the most prominent spatial features learned for the given input.}
    \label{fig:gabor_features}
\end{figure*}

Considering the \emph{Task 2} results, the comparison between the baseline models suggest that exploring not only the depth but also the width of the networks, i.e. by adding parallel operations to their branches, improves performance. In this context, the GSE ResNeXt model achieved a 2.58\% higher MCC than the Xception model, which obtained the second-best result. This demonstrates that combining attention and Gabor convolutional layers is also beneficial for this classification problem. 
The Gabor convolutional layer, the first layer in the GSE ResNeXt architecture, approximates a two-dimensional band-pass filter. Consequently, the features extracted after this layer represent the various two-dimensional frequency bands present in the input spectrogram, as illustrated in \figureref{fig:gabor_features}, which shows the feature maps obtained for an example spectrogram. It is possible to notice how these maps exhibit diverse textures across channels. Applying channel attention after these features introduces a frequency-selection behaviour, focusing the training towards the relevant spatial frequencies in the input representation and thereby enhancing feature discrimination, noise robustness, and convergence.

\begin{figure}[!bht]
    \centering
    \resizebox{\columnwidth}{!}{\includegraphics{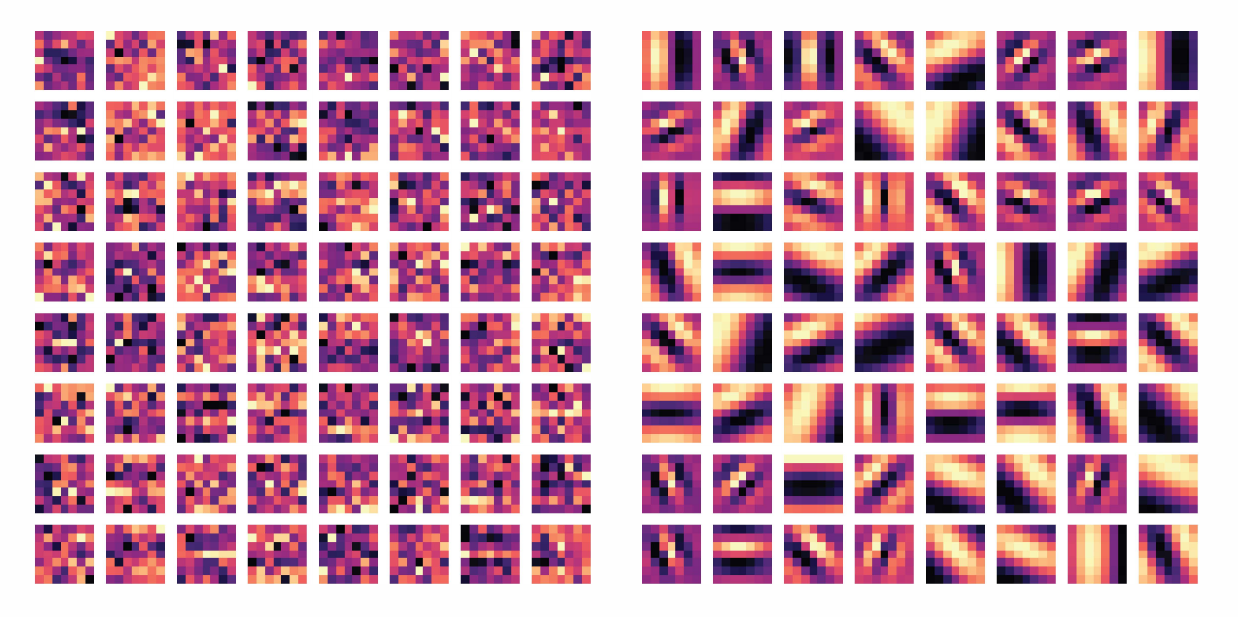}}
    \caption{A comparison of the first-layer kernels obtained from a ResNeXt model using traditional convolutions (left) and Gabor convolutions (right). Traditional convolutional kernels tend to converge to complex and potentially noisy representations, whereas Gabor kernels emphasise spatial relationships.}
    \label{fig:conv_kernels}
\end{figure}

To evaluate the noise robustness of the Gabor convolutional layers, an analysis was conducted on the initial kernels learned by ResNeXt and GSE ResNeXt models. As shown in \figureref{fig:conv_kernels}, the kernels learned using traditional convolutions converged to a sparser representation with greater variation in magnitude between neighbouring pixels. Conversely, Gabor convolutional kernels naturally preserve spatial frequencies, contributing to a improved noise robustness. The convergence of Gabor kernels to a diverse range of natural frequencies and orientations contributed to a more comprehensive representation of the input data, which also influenced the final classification accuracy.

\begin{figure}[!bht]
    \centering
    \resizebox{\columnwidth}{!}{\includegraphics{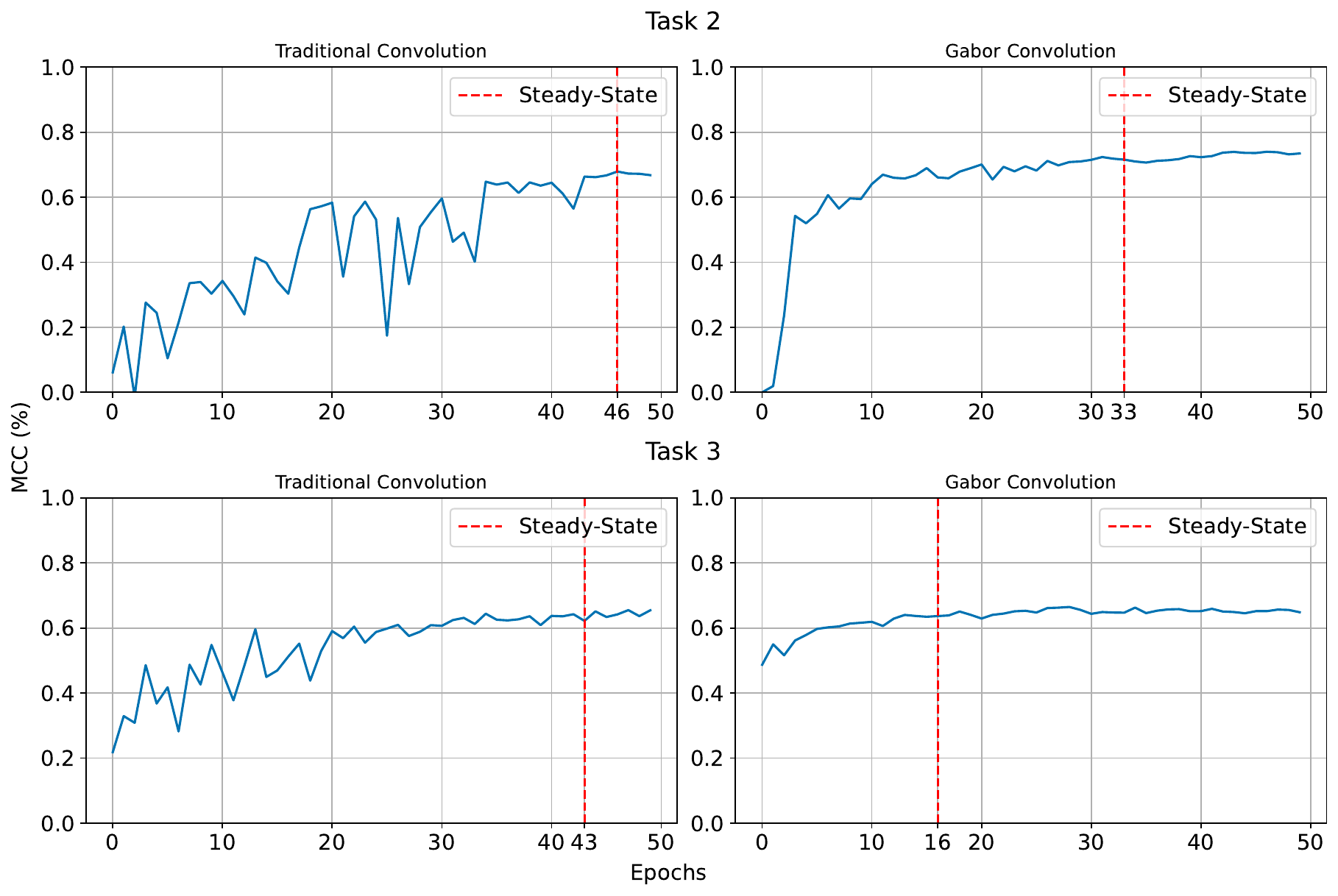}}
    \caption{Comparison of the MCC values obtained on the validation dataset during the training stage of a SE ResNeXt model with traditional convolutions and a SE ResNeXt model with Gabor convolutions in the first layer. The red dashed line indicates the epoch at which training reached a steady state. The use of Gabor convolutions resulted in a more stable validation metric and faster convergence compared to traditional convolutions.}
    \label{fig:mcc_stability}
\end{figure}

To evaluate the training convergence, the epoch at which the training achieved a steady state was identified. In this experiment, the steady state was defined as the epoch at which the validation MCC was within 2\% of the final value for three consecutive epochs. Using Gabor convolutions in the first layer, the steady state was achieved at epoch 33 for \emph{Task 2}; however, it was only achieved at epoch 46 when traditional convolutions were used. This represents a 28\% improvement, as can be seen in \figureref{fig:mcc_stability}. The Gabor convolution layer significantly improved the convergence of the ResNeXt model, also improving the final performance. It is important to note that the improvements observed for \emph{Task 2} were also observed for \emph{Task 3}, where the final state was reached at epoch 16 with Gabor convolutions compared to 43 with traditional ones, an improvement of 62\%, further supported by the superior classification performance of the GSE ResNeXt model. Nothing can be said about \emph{Task 1}, since the classification accuracy was almost perfect in all baseline models.

At the same time that \emph{Task 2} represented a middle ground between tasks 1 and 3, it also enables discussion about the temporal proximity of the data. As \emph{Task 2} focuses on classifying the same target in a different position on the time axis, evaluating the behaviour of the classification algorithm with respect to this particular property can provide a better understanding of the model's overall performance. Additionally, an evaluation can also be conducted regarding the size of the dataset needed to train the models, aiming to enable the application of these pipelines in scenarios involving limited amounts of data.

\subsection{Temporal proximity and dataset size evaluation}
\label{subsec:additional_task2}

\begin{figure}[!bht]
    \centering
    \resizebox{\columnwidth}{!}{\includegraphics{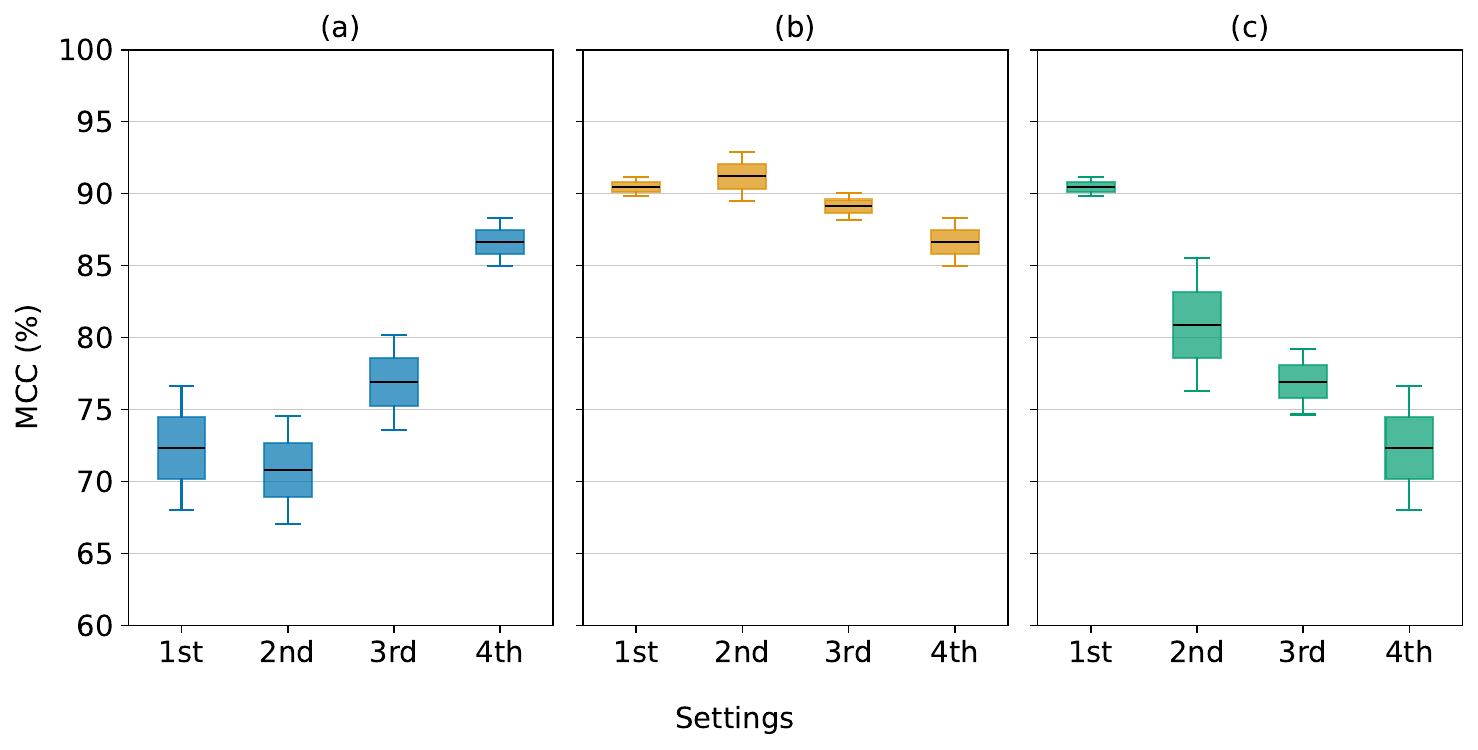}}
    \caption{Classification metrics obtained for the \emph{Task 2} experiments (a), (b), and (c), as illustrated in \figureref{fig:task2_exp}. Experiment (a) demonstrates improved performance when the temporal distance is shorter; experiment (b) shows stable classification performance regardless of dataset size; and experiment (c) combines both observations, with decreasing performance as temporal distance increases.}
    \label{fig:task2_exp_results}
\end{figure}

The results of the first experiment (a), which can be seen in \figureref{fig:task2_exp_results}, support the idea that adjacent segments share sufficient similarities to facilitate classification. When the segments become more distant, the classification deteriorates, with the worst-case scenario showing a difference of 14 MCC percentage points. As the recording represents a moving target, performance variation is directly related to the change in physical distance between target and sensor, since the same target at different distances will not sound the same.
Factors such as frequency attenuation and multi-path sound wave reflections can distort the perception of the received sound, thereby increasing the misclassification rates \parencite{stojanovic2008underwater}. This result raises an important concern about the performance of \emph{Task 3}. Since the classification performance is inaccurate when the recordings of the same vessels are separated in the temporal axis, the classification performance for different vessels will also be inaccurate, as is the case for \emph{Task 3}. Additional improvements in signal processing could enhance the similarity of recordings produced under different environmental conditions and, consequently, improve the final classification of both \emph{Task 2} and \emph{Task 3}.

Experiment (b), shows promising results concerning classification with reduced amount of data, as shown in \figureref{fig:task2_exp_results}. When both subsets are close to each other in terms of the temporal axis, the variation in training dataset size did not represent a significant impact on the final performance of the model, with a variation of less than 4 percentage points in MCC by using a four times smaller dataset. This conclusion suggests that classification performance may depend less on the quantity of training data and more on the similarity between the training representations and the unseen data distribution. 

The results of experiment (c), reinforce the conclusions obtained in experiments (a) and (b). They show that classification becomes increasing inaccurate when the training and test datasets are recorded under increasingly different conditions, in this case the distance between the target and the sensor. Once again, a robust pre-processing strategy that mitigates the impact of these conditions could improve the accuracy of the systems.

\subsection{Ablation Studies}
\label{subsec:ablation}

A series of ablation studies were conducted to evaluate the contribution of individual components within the GSE ResNeXt model, highlighting how classification performance is affected by the removal of each block. 
To isolate the impact of the two core components that define the novelty of this approach, the Gabor module and the SE mechanism, the analysis focused on their incremental contribution. The remaining parts of the architecture follow standard ResNeXt design choices, which have been extensively validated in prior literature \parencite{xie2017aggregated}.
The results, shown in \tableref{table:ablation_studies}, demonstrate the effectiveness of incorporating these components into the original architecture for tasks 2 and 3.

\begin{table*}[bht]
\centering
\resizebox{\textwidth}{!}{
    \begin{tabular}{lcccc}
    \hline
    \multirow{2}[1]{*}{\textbf{Model}} & \multicolumn{2}{c}{\textbf{MCC (\%)}} & \multicolumn{2}{c}{\textbf{Accuracy (\%)}}\\
	   & \textbf{Task 2} & \textbf{Task 3} & \textbf{Task 2} & \textbf{Task 3}\\
    
    \hline
    GSE ResNeXt                         & 78.53 $\pm$ 2.50 & 64.86 $\pm$ 4.83 & 83.85 $\pm$ 1.85 & 73.31 $\pm$ 3.75\\
       - SE layer                       & 78.03 $\pm$ 2.22 & 61.86 $\pm$ 3.18 & 83.49 $\pm$ 1.66 & 71.17 $\pm$ 2.40\\
       - Gabor Convolution              & 74.57 $\pm$ 2.73 & 65.46 $\pm$ 4.91 & 80.88 $\pm$ 2.03 & 73.89 $\pm$ 3.91\\
       - Gabor Conv, SE layer (ResNeXt) & 76.83 $\pm$ 0.24 & 53.00 $\pm$ 4.86 & 82.56 $\pm$ 0.22 & 64.09 $\pm$ 4.36\\
    \hline
    \end{tabular}
}
\caption{Ablation studies performed with the GSE ResNeXt model for tasks 2 and 3. Data is provided as mean $\pm$ standard deviation across folds. First line represents the results using the complete model and subsequent lines represents the removal of each block.}
\label{table:ablation_studies}
\end{table*}

In \emph{Task 2}, performance was significantly impacted by the removal of the Gabor convolutional layer. The configuration using SE layers with traditional convolutions yielded the lowest performance. Since SE attention operates on channel-wise information, its effectiveness depends on the quality of channel representations produced by the model. With traditional convolutions, the resulting channel representations were noisy, as shown in \figureref{fig:conv_kernels}, leading to reduced classification performance. Removing both Gabor convolutions and SE attention produced less variation across folds, however, resulted in a substantial reduction in the final classification coefficient compared to configurations that included Gabor convolutions.

For \emph{Task 3}, interpretation is limited by the considerable standard deviation between folds, which represents up to 9\% of the average MCC. Still, the removal of both channel attention and Gabor convolutions affected the classification performance, with a reduction of 18\% in MCC. Although removing Gabor convolutions achieved an average MCC about 0.6 percentage points higher than GSE ResNeXt, its performance varied greatly across folds, with a standard deviation of 4.91, making the difference not practically significant. Moreover, traditional convolutions showed slower and less stable convergence than Gabor convolutions, as illustrated in \figureref{fig:mcc_stability}, reinforcing their importance in the final model.

\section{Final Remarks}
\label{sec:conclusions}
This research presents the GSE ResNeXt model, a residual neural network that incorporates Gabor convolutions in the initial layer and applies the squeeze-and-excitation attention mechanism in the network's backbone. The combination of these two components functions as a two-dimensional frequency selection mechanism, enhancing both noise robustness and model convergence. 
The model's performance in underwater acoustic classification was evaluated across three tasks of increasing difficulty, offering a comprehensive performance evaluation. The classification results demonstrate the superiority of the GSE ResNeXt model over commonly used architectures such as Xception and MobileNet V2, improving classification MCC by 2.58\% and reducing training time by 28\% when considering the scope of \emph{Task 2} and 62\% for \emph{Task 3}.

The in-depth analysis of \emph{Task 2}, which focuses on the temporal dependency of the recordings, provided a consistent evaluation of the results. Unlike previous studies, which primarily address the outcomes of \emph{Tasks 1} and \emph{3}, this work presents an evaluation of the impact of temporal distance between the training and testing datasets. Since the recordings represent moving targets, variations in temporal distance are directly associated with changes in physical distance. The results indicate that classification performance deteriorates proportionally with increasing distance between subsets, even when the same vessel is observed in different time windows. This finding also suggests that improvements in the data processing stage may enhance performance by potentially mitigating environmental factors that alter a ship’s radiated noise over distance. Furthermore, the results imply that the amount of training data is less influential than distance in determining classification performance, indicating that future research should prioritise signal processing techniques over enhancements to classification models.

When assessing the impact of both Gabor convolutional layers and attention mechanisms on the results of \emph{Task 2}, their contribution is evident, with a 2.2\% improvement in classification performance, and even more pronounced for \emph{Task 3}, which shows a 22.3\% gain. These results highlight the effectiveness of the frequency selection mechanism in classifying underwater acoustic targets, positioning it as a strong alternative for reducing training time and improving classification performance.

Despite the improved classification performance, the proposed approach remains relatively heavy for embedded applications compared to lightweight models such as MobileNet V2 and MobileViT. This opens opportunities for compression or pruning to reduce computational cost. Efficiency could also benefit from reducing the input frequency range through resampling or band filtering, eliminating redundancies in spectral data. Subsequent steps in this research could focus on mitigating frequency attenuation inherent to the underwater acoustic domain, assessing the impact of multi-path sound wave reflections on acoustic perception, and performing cross-dataset validation by adapting the evaluation framework for smaller datasets.

\section*{Acknowledgements}
This work was supported by the National Industry PhD Program, Australia, in a partnership between Flinders University and PrioriAnalytica Pty Ltd.

\bibliographystyle{elsarticle-harv} 
\bibliography{references}

\end{document}